%% file: main.tex
\documentclass[conference]{IEEEtran}
\usepackage{cite}
\usepackage{amsmath,amssymb,amsfonts}
\usepackage{algorithmic}
\usepackage{graphicx}
\usepackage{textcomp}
\usepackage{xcolor}
\usepackage{fancyhdr}
\usepackage[hyphens]{url}
\usepackage{blindtext}
\usepackage{caption}
\usepackage{subcaption}
\usepackage{hyperref}

\def\BibTeX{{\rm B\kern-.05em{\sc i\kern-.025em b}\kern-.08em
    T\kern-.1667em\lower.7ex\hbox{E}\kern-.125emX}}

\makeatletter
\newcommand*\titleheader[1]{\gdef\@titleheader{#1}}
\AtBeginDocument{%
  \let\st@red@title\@title
  \def\@title{%
    \bgroup\normalfont\large\centering\@titleheader\par\egroup
    \vskip0.5em\st@red@title}
}
\makeatother
\newcommand{\hpcasubmissionnumber}{xxx}
\hypersetup{
    colorlinks=true,
    linkcolor=blue,
    filecolor=magenta,      
    urlcolor=cyan,
    citecolor = magenta,
    pdfpagemode=FullScreen,
}
\fancypagestyle{firstpage}{
  \fancyhf{}

  \fancyhead[C]{\normalsize{HPCA 2022 Submission
      \textbf{\#\hpcasubmissionnumber} \\ Confidential Draft: DO NOT DISTRIBUTE}} 
  \fancyfoot[C]{\thepage}
}

\title{Real-Time Fully Unsupervised Domain Adaptation for Lane Detection in Autonomous Driving\vspace{-0.1in}}

\titleheader{\vspace{-0.25in} \normalsize 2023 Design, Automation \& Test in Europe Conference (DATE 2023) - Late Breaking Results}
\author{Kshitij Bhardwaj$^{*}$, Zishen Wan$^{\dag}$, Arijit Raychowdhury$^{\dag}$, Ryan Goldhahn$^{*}$ \\ \normalsize \textit{$^{*}$Lawrence Livermore National Laboratory \hspace{0.05in} $^{\dag}$Georgia Institute of Technology} \vspace{-0.16in}}

\begin{document}

\maketitle
\pagestyle{plain}

%%%%%% -- PAPER CONTENT STARTS-- %%%%%%%%

\begin{abstract}

While deep neural networks are being utilized heavily for autonomous driving, they need to be adapted to new unseen environmental conditions for which they were not trained. We focus on a safety critical application of lane detection, and propose a lightweight, fully unsupervised, real-time adaptation approach that only adapts the batch-normalization parameters of the model. We demonstrate that our technique can perform inference, followed by on-device adaptation, under a tight constraint of 30 FPS on Nvidia Jetson Orin. It shows similar accuracy (avg. of 92.19\%) as a state-of-the-art semi-supervised adaptation algorithm but which does not support real-time adaptation.

\end{abstract}

\input{intro}
\input{benchmarks}
\input{approach}
\input{results}

\input{conclusion}
\input{acknowledgement}

%%%%%%% -- PAPER CONTENT ENDS -- %%%%%%%%

%%%%%%%%% -- BIB STYLE AND FILE -- %%%%%%%%
%\bibliographystyle{IEEEtranS}
\bibliographystyle{unsrt}
\bibliography{refs}
%%%%%%%%%%%%%%%%%%%%%%%%%%%%%%%%%%%%

\end{document}

%% file: intro.tex
% \vspace{-0.05in}
\section{Introduction}
\label{sec:intro}
%\vspace{-0.05in}

Since deep neural networks (DNNs) are being used for safety-critical actions in autonomous driving, it is important that the deployed DNNs are robust to noise and can quickly adapt to environmental changes for which they have not been trained. These models are typically trained using simulators (e.g., CARLA~\cite{dosovitskiy2017carla}). However, the training data (source domain) can be significantly different from real-world conditions (target domain). 
%We refer to the former as the source domain and the latter as the target domain. 
Deep learning models will need to be adapted from the labeled source domain to the unlabeled target domain (known as Unsupervised Domain Adaptation or UDA).
%Since deep neural networks (DNNs) are being used for safety critical actions in autonomous driving, it is important that the deployed DNNs are robust to noise and are also able to quickly adapt to changes in their environment for which they have not been trained. These models are typically trained using simulators such as CARLA~\cite{dosovitskiy2017carla}. However, the training data can be significantly different from the real-world conditions. We refer to the former as the source domain and the latter as the target domain. Deep learning models will need to be adapted from the labeled source domain to the unlabeled target domain (known as Unsupervised Domain Adaptation or UDA).

Adaptation of neural networks is typically performed through cloud, which may not be feasible or preferred. Such adaptation requires the vehicle to be connected to the cloud which is not always the case, e.g., remote locations and choppy connectivity. Additionally, we may not want to share data due to privacy concerns. Furthermore, while the model adapts, the conditions might again change before the updated model is deployed from the cloud, in which case its accuracy will suffer. Ideally, the deployed model should adapt on device and in real-time as it sees new data. The challenge is that the new data will not be labeled and the autonomous vehicle may not have enough computing resources to rapidly re-train the model on the new data in real time and still meet all the deadlines. Therefore, there is a need for a very lightweight unsupervised DNN adaptation approach for constrained computing platforms.

A recent work introduced CARLANE benchmark and a sim-to-real domain adaptation technique for 2D lane detection~\cite{stuhrcarlane}. Lane detection is an extremely important step that must be highly robust for the safety of the autonomous vehicle. While this work focused on a critical application, there are several limitations: (i) adaptation is not memory-efficient as it requires the use of both labeled source data (that the model is initially trained with) and unlabeled target data; (ii) the approach cannot be used in real time on a low-power computing chip as it runs for 10s of epochs and uses several thousands of source and training data samples for training, which causes significant latency/energy overheads; and (iii) it also generates pseudo labels for the target data, which incurs extra overheads.

In contrast, we propose a lightweight, fully unsupervised and real-time DNN adaptation algorithm for lane detection. Our approach updates only the batch-normalization layers' hyperparameters based only on unlabeled target data. We demonstrate this technique on Nvidia Jetson Orin, where we show that inference, followed by model adaptation, using each incoming $1280\times720$ image can be achieved in tight real-time performance constraints of up to 30 FPS. The updated model is then used for the next image. Our online adaptation approach achieves almost similar lane detection accuracy for CARLANE benchmarks as the current state-of-the-art adaptation algorithm~\cite{stuhrcarlane} but without the use of extra labeled data and can be performed in real time. Our code will be open-sourced.

\begin{figure}[t]
\centering
  \includegraphics[width=1\columnwidth]{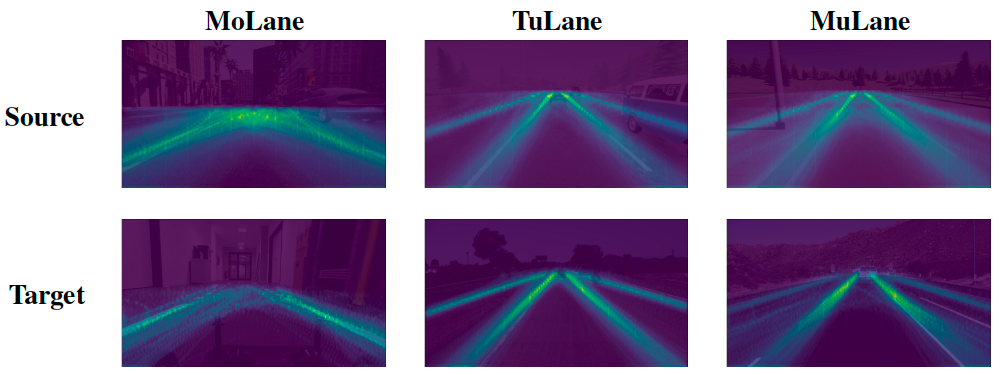}\vspace{-0.05in}
  \caption{CARLANE adaptation benchmarks~\cite{stuhrcarlane}.}\vspace{-0.2in}
  \label{fig:benchmarks}
\end{figure}

%% file: benchmarks.tex
% \vspace{-0.05in}
\section{Lane detection adaptation baseline}
%\vspace{-0.05in}
%\label{sec:dse}

We use the ultra fast lane detection (UFLD) algorithm~\cite{qin2020ultra}. 
%The approach is more computationally efficient than typical deep segmentation techniques. This method formulates lane detection problem as a row-based selection based on global image features. 
%It selects the correct locations of lanes on each predefined row using global features.
% instead of segmenting and classifying each pixel.
Lanes are represented as a series of horizontal locations at predefined rows (row anchors). On each row anchor, the location is divided into many grid cells. Therefore, the detection of lanes is formulated as selecting correct cells over row anchors.

Figure~\ref{fig:benchmarks} shows the three types of CARLANE benchmarks used: MoLane (2 lanes), TuLane (4 lanes), and MuLane (4 lanes). The source data for these benchmarks are taken from CARLA simulations on which the DNNs are initially trained using the UFLD algorithm. The unlabeled target or testing data, on which the domain adaptation of the pre-trained DNNs is performed, are: (i) real-world model vehicle data for MoLane; (ii) TuSimple real-world images of U.S. highways dataset~\cite{tusimple} for TuLane; and (iii) both model vehicle data and TuSimple data for MuLane (i.e., a multi-target benchmark). The images are $1280\times720$, collected using a 30 FPS camera.

We use a state-of-the-art domain adaptation technique for lane detection as the baseline~\cite{stuhrcarlane}. It adapts the UFLD-trained models by: (i) encoding the semantic structure of data in both the source target domains into an embedding space. K-means is used for this encoding; (ii) transferring knowledge from source to target using the embeddings; and (iii) updating all of the DNN parameters using backpropagation. 

While the baseline achieves excellent accuracy, it is not suitable for real-time adaptation due to several high overhead steps (learning embeddings, knowledge transfer, K-means, and re-training the pre-trained model for 10 epochs). It also requires a significant amount of labeled source data on device, which adds extra memory/data transfer overheads. Each epoch on Orin took greater than 1 hour (depending on the benchmark) hence, making it unsuitable for real-time adaptation.

%% file: approach.tex
% \vspace{-0.05in}
\section{Real-time Lightweight Adaptation Technique}
\label{sec:approach}
% \vspace{-0.05in}

Our approach, called {\em LD-BN-ADAPT,} updates only the batch-normalization (BN) parameters of the deployed UFLD model (pre-trained using the source data). BN parameters typically only comprise of 1\% of the total model parameters, hence updating these parameters is lightweight. In real time, adaptation is performed, using a batch of collected unlabeled target data, right after performing inference. In particular, each BN layer performs two steps: (i) normalization that standardizes the input $x$ into $x' = (x-\mu)/\sigma$ using its mean and standard deviation, and (ii) transformation that turns $x'$ into $x'' = \gamma x' + \beta$ using scale ($\gamma$) and shift ($\beta$) parameters. During adaptation, while (i) are recomputed from the unlabeled data, scale and shift parameters in (ii) are optimized by a loss function while running a {\em single} backpropagation pass. Since the optimization is performed using only unlabeled data, {\em entropy of model predictions} is used as the loss function. Shannon entropy for a prediction $y$ is defined as: $H(y) = - \sum_{c} p(y_c)log p(y_c)$ for probability of $y$ for class $c$ (dimensions of $y$ are $grid_{cells} \times row_{anchors} \times num_{lanes}$, where $grid_{cells}$ is 100, $row_{anchors}$ is 56 and $num_{lanes}$ can be 2 or 4 depending on the benchmark).
%In our study, we also consider unsupervised adaptation of other types of parameters as well such as of convolutional layers (LD-Conv-Adapt) and fully connected layers (LD-FC-Adapt), and compare them with LD-BN-Adapt. For both the former adaptations also, we use Shannon entropy as the loss function. 
While a previous work has looked at BN-based UDA~\cite{wang2020tent}, they have focused on image classification and did not target on-device real-time adaptation, while we focus on much more complex lane detection task with strict real-time deadlines. In addition to BN-based adaptation, we also tested convolutional and fully-connected adaptation but found the BN-based approach to be the most effective. Pytorch-1.11 is used for implementation.

%% file: results.tex
% \vspace{-0.1in}
\section{Measurement Results and Analysis}
\label{sec:results}
% \vspace{-0.05in}

\begin{figure}[t]
\centering
  \includegraphics[width=1\columnwidth,trim=4 4 4 4,clip]{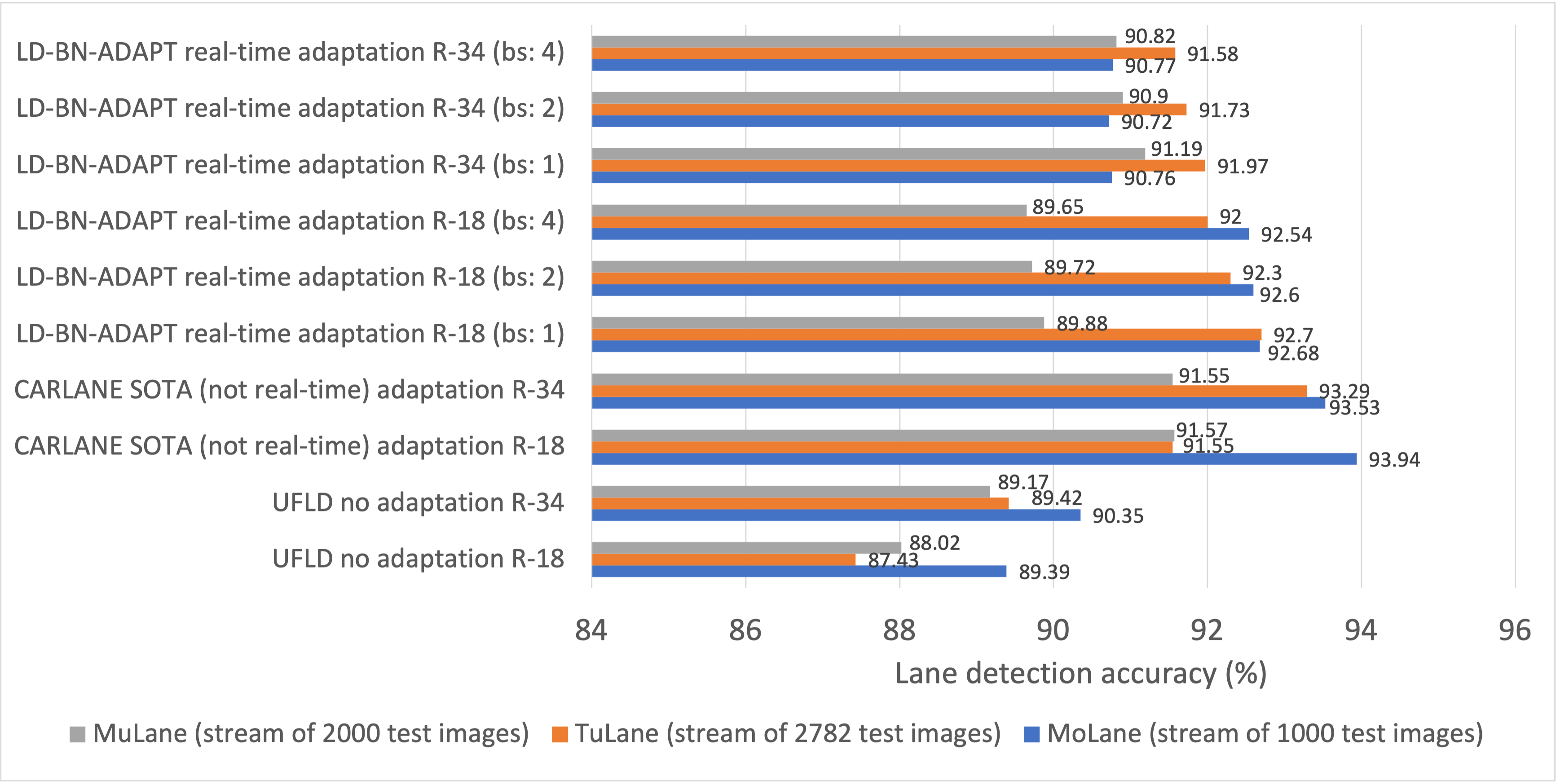}\vspace{-0.05in}
  \caption{Lane detection accuracy results.}
  \label{fig:acc}
\end{figure}

Figure~\ref{fig:acc} shows the lane detection accuracy for: (i) UFLD no adaptation; (ii) CARLANE SOTA adaptation technique which is not a real-time (or test-time) approach; (iii) real-time LD-BN-ADAPT with varying batch sizes (bs) of 1, 2, 4 (i.e., adaptation after every image or 2/4 images); and (iv) two types of ResNet models (ResNet-18/R-18, ResNet-34/R-34). For both the models, LD-BN-ADAPT with batch size 1 shows the best accuracy compared to other batch sizes, and is very close to the CARLANE SOTA. CARLANE SOTA's best accuracies for MoLane, TuLane, and MuLane are: 93.94\% (R-18), 93.29\% (R-34), 91.57\% (R-18), respectively {\em (avg. of 92.93\%),} and LD-BN-ADAPT's are: 92.68\% (R-18), 92.7\% (R-18), and 91.19\% (R-34) {\em (avg. of 92.19\%).}

\begin{figure}[t]
\centering
  \includegraphics[width=.9\columnwidth,trim=4 4 4 4,clip]{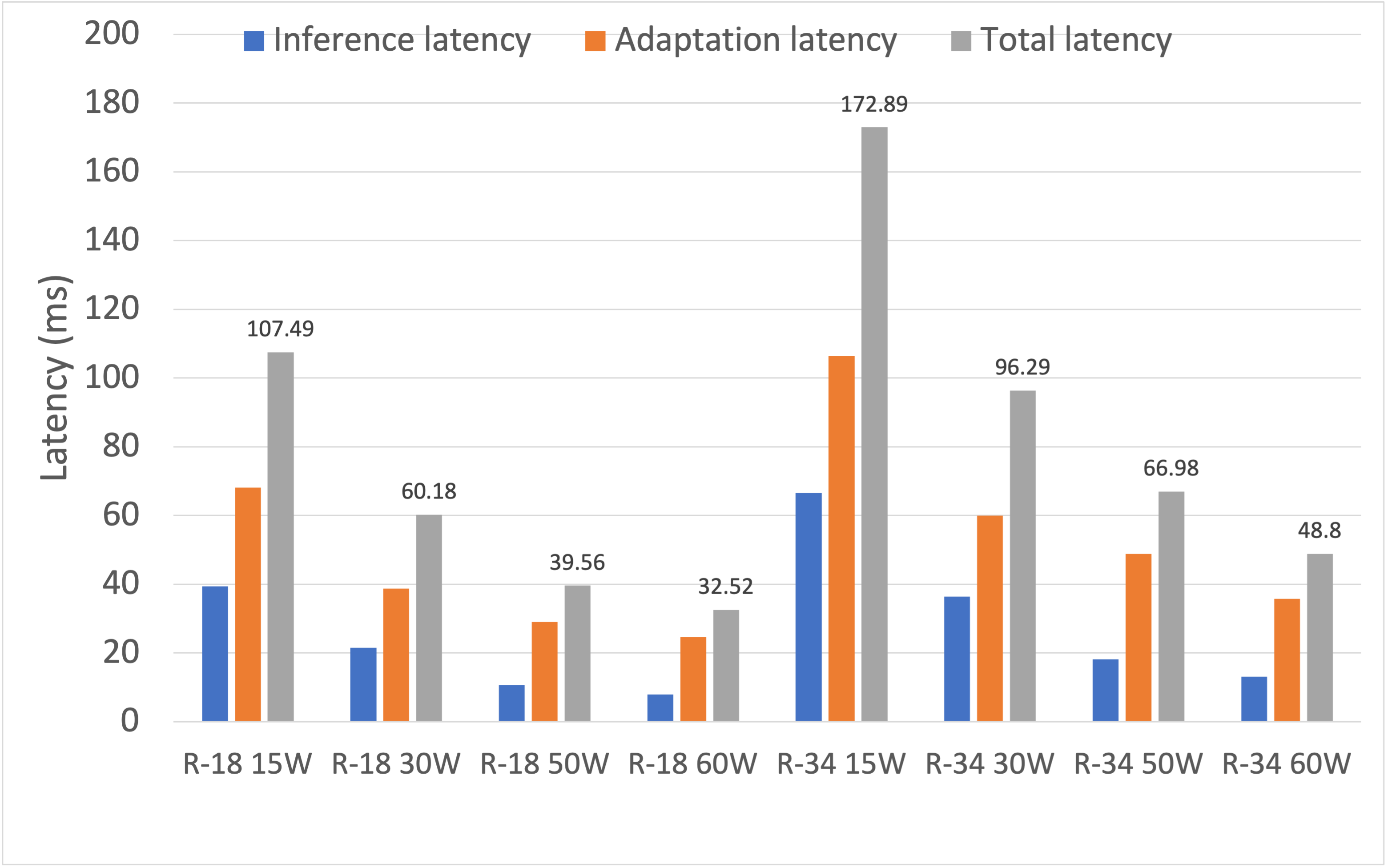}\vspace{-0.15in}
  \caption{Lane detection latency on Nvidia Jetson Orin for various power modes for LD-BN-ADAPT (batch size: 1).}\vspace{-0.1in}
  \label{fig:lat}
\end{figure}

Figure~\ref{fig:lat} shows the latency results on Nvidia Jetson Orin, for its different power modes, for LD-BN-ADAPT with batch size 1 (other batch sizes not considered as they show lower accuracy). Using R-18 with 60W power mode meets the strict real-time constraint of 30 FPS (i.e., 33.3 ms deadline), demonstrating that inference followed by model adaptation is possible in real time even with such tight constraints. For more relaxed constraints, such as 18 FPS or a deadline of 55.5 ms (similar to Audi A8 sedan with level 3 autonomous driving system~\cite{visionArticle}), R-18 at 60W, R-18 at 50W, and R-34 at 60W meet the constraint. In this case, the best model can be selected based on the power constraints and the type of task. For example, if there is a strict power constraint of 50W then R-18 should be used. On the other hand, if a more robust model is required that shows better accuracy for multi-target scenarios (e.g., MuLane) then R-34 should be selected. 

These results demonstrate that real-time model adaptation for a complex and safety-critical task, such as lane detection, is possible but requires a careful study of the multi-objective design space and the various application constraints.

%% file: conclusion.tex
% \vspace{-0.1in}
\section{Conclusion}
\label{sec:conclusion}
% \vspace{-0.05in}

We propose a real-time, lightweight, and fully unsupervised model adaptation approach for lane detection, which only adapts BN parameters of the pre-trained models. We demonstrate that inference, followed by our real-time adaptation, can meet the tight constraints of up to 30 FPS on Jetson Orin.

% \vspace{-0.1in}

%% file: acknowledgement.tex
\section*{Acknowledgement}
This work was performed under the auspices of the U.S. Department of Energy by Lawrence Livermore National Laboratory under Contract DEAC52-07NA27344 (IM: LLNL-CONF-843048). ZW and AR were supported in part by CoCoSys, one of the seven centers in JUMP 2.0, a Semiconductor Research Corporation (SRC) program sponsored by DARPA.